\documentclass[10pt,twocolumn,letterpaper]{article}

\usepackage[pagenumbers]{cvpr} 

\usepackage{stfloats}
\usepackage{booktabs}
\usepackage{graphicx}
\usepackage{multirow}
\usepackage{capt-of}
\usepackage{algorithm}
\usepackage{algorithmic}
\usepackage{xcolor}
\newcommand{\algcomment}[1]{\textcolor{gray}{// #1}}

\definecolor{cvprblue}{rgb}{0.21,0.49,0.74}
\usepackage[pagebackref,breaklinks,colorlinks,allcolors=cvprblue]{hyperref}

\def\paperID{5} 
\def\confName{CVPR}
\def\confYear{2026}

\title{DAMA: Disentangled Body-Anchored Gaussians for Controllable~Multi-Layered~Avatars}

\author{
Daniel Eskandar$^{1,2,5}$ \quad
Berna Kabadayi$^{1,3}$ \quad
Garvita Tiwari$^{1,2,4}$ \quad
Gerard Pons-Moll$^{1,2,4}$ \\ \\
{\small
$^{1}$University of Tübingen, Germany \quad
$^{2}$Tübingen AI Center, Germany \quad
$^{3}$Max Planck Institute for Intelligent Systems, Germany
}\\
{\small
$^{4}$Max Planck Institute for Informatics, Germany \quad
$^{5}$Zuse School ELIZA, Germany
}
}

\begin{document}

\twocolumn[{
\maketitle

\vspace{-12mm}
\begin{center}
\begin{minipage}{\textwidth}
\centering
\includegraphics[width=\textwidth]{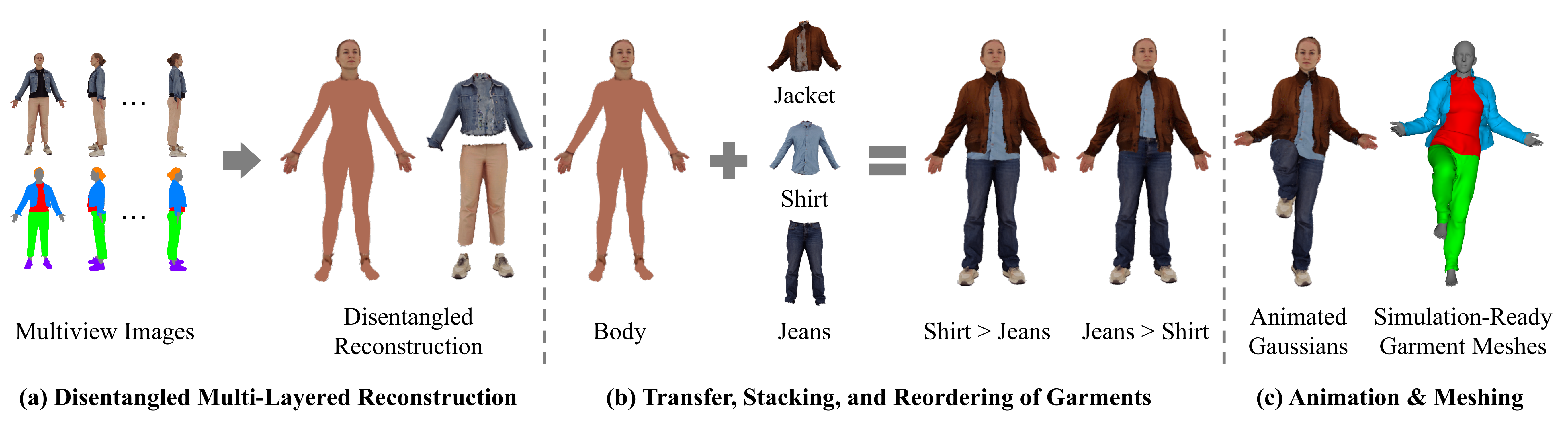}
\vspace{-5mm}
\captionof{figure}{We present \textbf{DAMA}, a method for reconstructing physically plausible multi-layered avatars. 
(a) From multi-view RGB images and masks, we reconstruct clean, intersection-free layers via body-anchored Gaussians. 
(b) The layers enable garment composition, stacking, and reordering (e.g., Shirt $>$ Jeans vs.\ Jeans $>$ Shirt). 
(c) The garments are animatable and convertible to simulation-ready meshes.}
\label{fig:teaser}
\end{minipage}
\end{center}
}]

\begin{abstract}

\vspace{-2mm}
Existing 3D clothed avatar reconstruction methods achieve high visual fidelity but ignore geometric structure and physical plausibility. They either model clothed humans as a single deformable surface or attempt garment disentanglement without enforcing geometric constraints, resulting in ambiguous garment boundaries and no control over stacking or layer ordering. To address these limitations, we introduce DAMA (\textbf{D}isentangled body-\textbf{A}nchored Gaussians for Controllable \textbf{M}ulti-layered \textbf{A}vatars), a 3D avatar reconstruction method that produces physically plausible clothed avatars through a dedicated representation and reconstruction method. At the representation level, we bind Gaussians to SMPL-X faces using barycentric in-plane coordinates and a positive normal offset. Based on this parameterization, the reconstruction method lifts 2D segmentations to body-anchored Gaussians, refines layers using topology-guided correction, and jointly optimizes geometry and appearance. DAMA is the first Gaussian avatar reconstruction method from multi-view images to achieve physically plausible layering, clean garment separation, and explicit stacking control. On the full 4D-DRESS dataset (82 scans), it achieves state-of-the-art performance in geometry reconstruction, garment separation, penetration rate, and penetration depth. The representation further supports user-defined garment reordering and fast conversion of body-conforming garments to simulation-ready meshes.
Project Page: \url{https://danieleskandar.github.io/dama/}

\end{abstract}

\vspace{-5mm}
\section{Introduction}

\vspace{-2mm}
Photorealistic 3D human avatars are essential for applications such as virtual reality and digital try-on \cite{xue2025infinihuman, he2025vton, kocabas2024hugs, li2024animatable}. A key challenge is modeling clothing, which is not a single surface but a composition of layered garments combined in different configurations. These garments remain in contact with the body and each other while preserving consistent ordering. Neural radiance fields \cite{mildenhall2020nerf, muller2022instant, weng2022humannerf, peng2021neural, peng2023implicit} and Gaussian-based representations \cite{kerbl3Dgaussians, huang20242dgs, li2024animatable, qian20243dgs, hu2024gaussianavatar, hu2024gauhuman} achieve high-fidelity reconstruction from images and videos, but prioritize rendering quality over explicit geometric and physical structure.

Previous works differ in how explicitly they model clothing structure. Single-surface avatar methods reconstruct the clothed human as a unified deformable geometry without garment decomposition \cite{feng2022capturing, feng2023learning, guo2024reloo}. They typically bind the geometry to a parametric body model such as SMPL-X \cite{pavlakos2019smplx} to drive articulation and animation. Another line of work models the body and clothing separately but merges all garments into a single layer \cite{feng2022capturing, feng2023learning}. This design enables whole-outfit transfer but does not provide per-garment control. Garment-level disentanglement has been explored in different ways. Some methods isolate a target garment from the remaining geometry \cite{kim2024gala, lin2024layga}. Others reconstruct the body and multiple garments as separate layers using template-based approaches \cite{pons2017clothcap, bhatnagar2019multi, jiang2020bcnet, tiwari2020sizer}. Recent Gaussian splatting methods also reconstruct the body and multiple garments as distinct layers from multi-view videos \cite{zielonka2025drivable, chen2026gaussianwardrobe}.

Many of these approaches rely on multi-frame optimization or video-based learning, tying disentanglement to temporal tracking rather than encoding geometric layer ordering \cite{lin2024layga, zielonka2025drivable, chen2026gaussianwardrobe}. On the other hand, Disco4D \cite{pang2025disco4d} infers the body and all garments jointly from image supervision alone. However, it enforces separation through optimization-based constraints, which often produce ambiguous boundaries and interpenetrating layers. Across these approaches, separation is not encoded in the representation, preventing consistent garment stacking and explicit layer control.

We introduce DAMA (\textbf{D}isentangled Body-\textbf{A}nchored Gaussians for Controllable \textbf{M}ulti-Layered \textbf{A}vatars). DAMA uses a novel Gaussian splatting representation that enforces layer ordering by design. Specifically, each Gaussian is anchored to a SMPL-X face by factorizing its mean into in-plane barycentric coordinates and a positive offset along the face normal. The barycentric parameterization binds each Gaussian to its assigned mesh face, preventing lateral drift to unrelated surface regions and preserves semantic identity under deformation. The positive normal offset constrains each garment layer to lie outwards along the surface normal, preventing interpenetration with the body and lower layers. This parameterization enforces layer ordering and intersection avoidance explicitly, unlike prior work that relies solely on optimization losses.

Leveraging this representation, we propose a novel reconstruction method that progressively optimizes geometry, segmentation, and appearance. First, we jointly reconstruct coarse geometry and segmentation by lifting 2D masks into a set of SMPL-X–anchored Gaussians. The lifted labels remain semantically aligned with the underlying mesh since the Gaussians are face-bound and cannot drift laterally. Then, we refine layer assignments using SMPL-X topology to correct inconsistent regions caused by occlusions or weak supervision. Finally, we refine geometry and texture for each garment under masked RGB supervision.

Our approach differs from prior work in three aspects. First, GALA \cite{kim2024gala} and Disco4D \cite{pang2025disco4d} lift 2D segmentations to 3D from a single frame, while LayGA \cite{lin2024layga} and Gaussian Wardrobe \cite{chen2026gaussianwardrobe} segment a first-frame template and supervise it with masks from video frames. Both strategies often produce noisy garment assignments. Our representation enables correcting the lifted labels. Gaussians remain semantically coupled to SMPL-X faces, enabling projection to the SMPL-X mesh and topology refinement for clean garment separation. Second, existing methods \cite{pang2025disco4d,kim2024gala,lin2024layga,chen2026gaussianwardrobe} optimize all layers jointly on the full image, whereas we optimize each garment independently. Third, prior work discourages intersections with penetration losses, while our positive normal offset enforces layer ordering by design and guarantees intersection-free reconstruction.

We evaluate DAMA on the full 4D-DRESS dataset (82 scans) \cite{wang20244ddress}, achieving state-of-the-art performance in geometry reconstruction, garment separation, penetration rate, and penetration depth while maintaining competitive rendering quality. The resulting avatars are fully animatable under SMPL-X articulation, enabling intersection-free motion. Beyond reconstruction accuracy, DAMA enables user-defined garment stacking and explicit layer ordering (e.g., selecting which garment lies over another). It also supports rapid conversion of body-conforming garments into simulation-ready meshes for downstream physical applications. We summarize our contributions as follows:

\begin{itemize}
    \item A novel parameterization for multi-layered avatars that binds Gaussian splats to SMPL-X faces with barycentric coordinates and a strictly positive normal offset.
    \item A topology-aware reconstruction method that progressively refines geometry, segmentation, and appearance.
    \item New clothing applications based on our representation: garment stacking and reordering, and fast conversion to simulation-ready garment meshes.
\end{itemize}


\label{sec:introduction}

\vspace{-2mm}
\section{Related Work}

\vspace{-2mm}
\noindent \textbf{Clothed Avatar Reconstruction.}
3D avatar reconstruction captures human appearance and motion. Early methods relied on parametric body models and mesh-based pipelines \cite{habermann2019livecap, habermann2020deepcap, xu2018monoperfcap, alldieck2018video, alldieck2019imghum, joo2018total, zhang2017detailed}, which support articulation and reposing but lack photorealistic detail. More recent methods use neural scene representations. Implicit neural fields \cite{mescheder2019occupancy, park2019deepsdf} and Neural Radiance Fields (NeRF) \cite{mildenhall2020nerf, muller2022instant} reconstruct avatars from monocular or multi-view video \cite{weng2022humannerf, jiang2022neuman, peng2021neural, peng2023implicit, peng2021animatable, jiang2023instantavatar} with high visual fidelity, but require slow per-scene optimization and costly mesh extraction. Explicit representations such as Gaussian Splatting \cite{kerbl3Dgaussians, huang20242dgs} enable faster training and real-time rendering. Several works adopt Gaussian-based models to reconstruct animatable avatars from monocular or multi-view video \cite{li2024animatable, qian20243dgs, hu2024gaussianavatar, hu2024gauhuman, shao2024splattingavatar, guo2023vid2avatar, jiang2025prioravatar, moon2024expressive, guo2025vid2avatar, kocabas2024hugs, lei2024gart, tan2025dressrecon}. However, these methods model clothed humans as a single fused surface. This supports appearance and pose control but lacks explicit garment decomposition and layer ordering, limiting garment-level manipulation and physical reasoning.

\noindent \textbf{Clothing Disentanglement.}
Many works separate clothing from the body. Some segment garments from 3D scans \cite{antic2024close, wang20244ddress, suzuki2025open} but do not extract animatable or transferable layers. Others use a two-layer representation (body layer and clothing layer) reconstructed from monocular video \cite{feng2022capturing, feng2023learning, guo2024reloo}. This enables whole-outfit transfer but fuses garments into one layer, preventing separate garment manipulation. A finer level of separation reconstructs garments as distinct layers. Some approaches recover multiple garments from 3D scans using predefined templates \cite{pons2017clothcap, tiwari2020sizer, bhatnagar2019multi, jiang2020bcnet}, requiring high-quality 3D input and limiting clothing diversity. Gaussian-based methods reconstruct garments from multi-view video \cite{lin2024layga, zielonka2025drivable} and jointly optimize segmentation and clothing deformation. This couples disentanglement with deformation, leaving layer order implicit rather than structurally encoded, which limits stacking and explicit layer control. Gaussian Wardrobe \cite{chen2026gaussianwardrobe} supports stacking but fixes layer order during training and resolves intersections after rendering, preventing reordering during inference. The closest works to ours are GALA \cite{kim2024gala} and Disco4D \cite{pang2025disco4d}, which focus on disentanglement during reconstruction rather than temporal deformation. GALA takes a single 3D scan, renders multi-view images, and separates one garment at a time using lifted 2D segmentations. Disco4D takes a single image, generates multi-view images, and jointly reconstructs body and garments. Both enforce separation through segmentation and penetration losses rather than encoding layer order in the representation, which leads to interpenetration, ambiguous garment boundaries, and no support for stacking or reordering.

\vspace{1mm}
\noindent \textbf{Clothed Avatar Generation.}
Generative methods synthesize clothed avatars from images or text prompts \cite{zhuang2024idolinstantphotorealistic3d, xue2024human3diffusion, sengupta2024diffhuman, dong2025iccv, shen2023xavatar, i_ho2024sith, saito2019pifu, xiu2022icon, Xiu2023econ, zhang2024sifu, liao2024tada, dong2024tela, lu2025gas}, often modeling body and clothing as a single surface. Some generate disentangled garments but only model geometry or do not support garment transfer \cite{corona2021smplicit, moon20223dclothed, wang2024disentangled, hu2025humanliff, aggarwal2022layered, Chen2024NeuralABC, vuran2025remu, li2022dig}. LayerAvatar \cite{zhang2025layeravatar} models texture and enables garment transfer, but restricts each garment type to one layer and does not support stacking or layer order control. Although these works address generation rather than reconstruction, they also do not encode geometric layer ordering. A representation with explicit stacking and layer ordering would enable greater control and physical consistency in generative settings.

\vspace{1mm}
\noindent \textbf{Body-Conditioned Representations.}
Prior work binds appearance or clothing to a parametric body model (e.g. SMPL-X) \cite{pavlakos2019smplx} to enable pose control and animation. Canonical NeRF methods learn a rest-space field warped to posed space via skeletal skinning and learned non-rigid offsets \cite{weng2022humannerf, jiang2022neuman, peng2021neural, peng2023implicit, peng2021animatable, jiang2023instantavatar}. Gaussian-based avatars predict canonical Gaussian maps on a template and deform the attached Gaussians with inherited linear blend skinning weights \cite{li2024animatable, lin2024layga, zubekhin2025giga}. D3GA \cite{zielonka2025drivable} embeds Gaussians in a tetrahedral cage that deforms with the body. GaussianAvatars \cite{qian2024gaussianavatars} represents each Gaussian mean in the local frame of a FLAME triangle \cite{li2017flame} and maps it to posed space through the animated mesh. Disco4D \cite{pang2025disco4d} applies the same triangle-local binding to clothing Gaussians on SMPL-X. SplattingAvatar \cite{shao2024splattingavatar} uses barycentric coordinates and a normal displacement but allows surface drift and bidirectional offsets while modeling the clothed human as a single deformable surface. These methods rely on losses to discourage drift and penetration but do not enforce geometric constraints. In contrast, we encode them directly: barycentric coordinates restrict Gaussians to their mesh face, and a strictly positive normal offset keeps layers outside the body.

\vspace{1mm}
\noindent \textbf{Clothing Applications.}
Clothed avatar applications range from generation \cite{xue2025infinihuman, zhang2025layeravatar, wang2025garmentcrafter}, animation and deformation modeling \cite{li2024animatable, shao2024splattingavatar, guo2024reloo, lin2024layga, rong2024gaussiangarments}, physical simulation \cite{grigorev2023hood, grigorev2024contourcraft, su2023caphy, li2024diffavatar, rong2024gaussiangarments, santesteban022snug, deluigi2023drapenet, li2022dig, kabadayi2026physheadsimulationreadygaussianhead}, relighting and cloth modeling \cite{saito2024rgca, wang2025rfbgca, guo2025pgc, peng2024pica, zheng2025physavatar}, to garment transfer or virtual try-on \cite{chen2024gaussianvton, ho2023learninglocally, han2018viton, sun2024outfitanyone}. Most methods focus on visual realism and assume a fixed or limited number of clothing layers. Gaussian Wardrobe \cite{chen2026gaussianwardrobe}, a recent concurrent work, supports stacking but enforces a predefined hierarchy (e.g., pants $<$ shirt $<$ jacket) and allows only one garment per layer. In contrast, our representation encodes explicit geometric layer order, enabling arbitrary garment stacking and user-defined reordering (e.g., shirt inside or outside pants).

\label{sec:related_work}

\vspace{-1mm}
\section{Method}

\vspace{-2mm}
\noindent \textbf{Preliminaries.}
Our method builds on SMPL-X~\cite{pavlakos2019smplx} for pose control and articulation, and Gaussian Splatting~\cite{kerbl3Dgaussians} as the explicit representation of the clothed avatar.

SMPL-X represents a human mesh as 
$M(\boldsymbol{\beta},\boldsymbol{\theta},\boldsymbol{\psi})=(\mathbf{V},\mathbf{F})$, 
where $\boldsymbol{\beta}$, $\boldsymbol{\theta}$, and $\boldsymbol{\psi}$ denote shape, pose, and expression parameters. 
The vertices $\mathbf{V}\in\mathbb{R}^{N_{\text{vertices}}\times 3}$ and faces $\mathbf{F}\in\mathbb{N}^{N_{\text{faces}}\times 3}$ have fixed topology under linear blend skinning (LBS). 
In DAMA, this surface is not rendered; instead, it anchors and deforms Gaussian layers.

Gaussian Splatting represents a scene as anisotropic Gaussians
$\mathcal{G}=\{g_i\}_{i=1}^{N_{\text{gaussians}}}$,
where each Gaussian
$g_i=(\boldsymbol{\mu}_i,\mathbf{s}_i,\mathbf{q}_i,\alpha_i,\mathbf{c}_i)$
has mean $\boldsymbol{\mu}_i$, scale $\mathbf{s}_i$, rotation $\mathbf{q}_i$, opacity $\alpha_i$, and colors $\mathbf{c}_i$, rendered via differentiable splatting and alpha compositing.
DAMA models the clothed human as layered Gaussian sets $\mathcal{G}^l$ (skin, hair, and garments). 
We adopt 2D Gaussian Splatting (2DGS) \cite{huang20242dgs}, which represents Gaussians as surface-aligned disks instead of volumetric blobs, enabling more stable surface modeling.

\begin{figure*}[t]
    \centering
    \includegraphics[width=\textwidth]{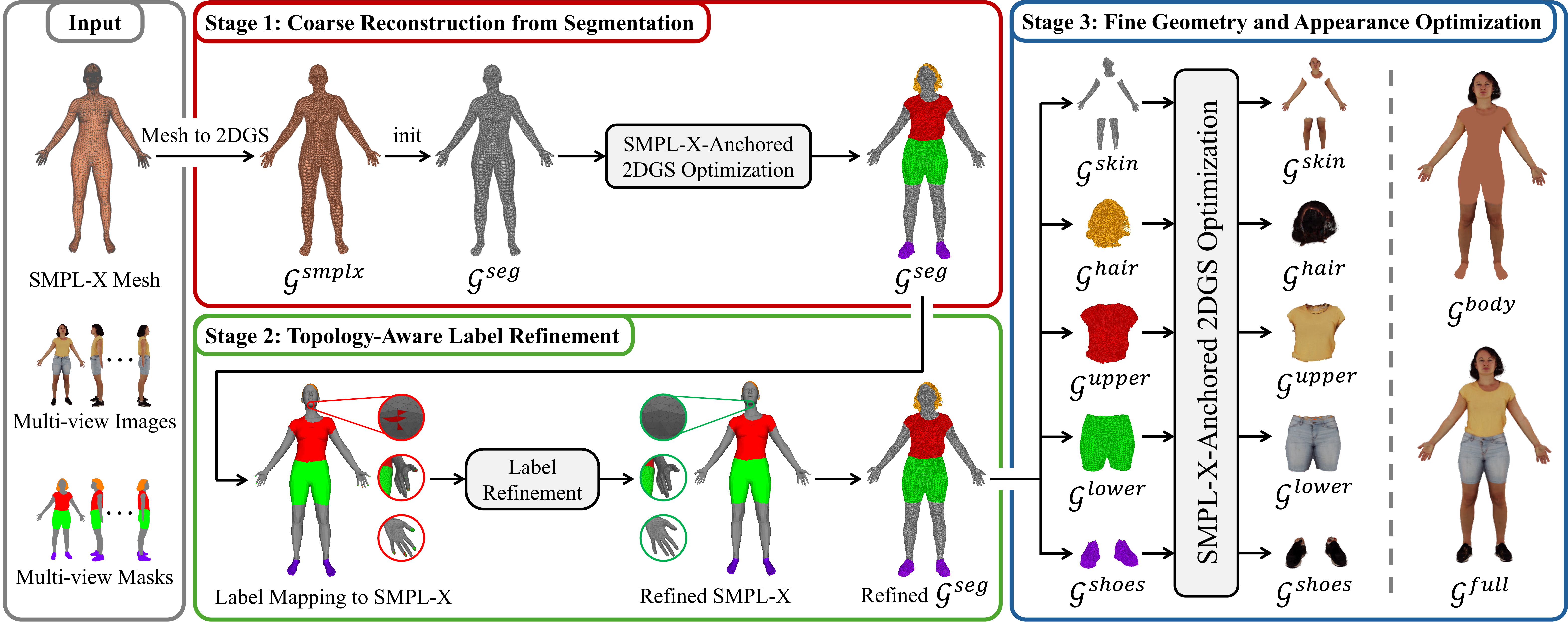}
    \vspace{-5mm}
    \caption{\textbf{DAMA Overview.} Given multi-view images and masks, we reconstruct a layered avatar with clean garment separation and no interpenetration. The method consists of three stages: (1) lifting 2D masks to SMPL-X–anchored Gaussians by optimizing coarse geometry and labels; (2) mapping labels to SMPL-X and refining them using mesh topology; (3) jointly optimizing geometry and appearance for each layer under masked RGB supervision. The final avatar guarantees clean disentanglement and intersection-free layering.}
    \label{fig:method}
    \vspace{-3mm}
\end{figure*}

\vspace{1mm}
\noindent \textbf{Problem Definition.}
Given multi-view RGB images $I_j$, segmentation masks $S_j$, camera intrinsics $K_j$, and extrinsics $T_j$ for $j=1,\dots,N_{\text{views}}$, and a fitted SMPL-X body, the goal is to reconstruct a layered Gaussian avatar. The avatar is represented as semantic Gaussian sets $\mathcal{G}^{l}=\{g_i\}_{i=1}^{N_l}$, each corresponding to a layer $l$ (e.g., skin, hair, shoes, or garments such as upper, lower, or outer clothing). The layers must remain cleanly separated and free of interpenetration with the body and with each other, while supporting SMPL-X–driven animation and garment transfer or stacking.

\vspace{2mm}
\noindent \textbf{Method Overview.}
The pipeline (Sec.~\ref{sec:pipeline}, Fig.~\ref{fig:method}) reconstructs a layered Gaussian avatar from multi-view images and masks in three stages: 
(1) lifting 2D segmentations to anchored Gaussians and optimizing coarse geometry and labels (Sec.~\ref{sec:segmentation}); 
(2) projecting labels to the SMPL-X mesh and refining via mesh topology (Sec.~\ref{sec:refinement}); 
and (3) jointly optimizing per-layer geometry and appearance under masked RGB supervision (Sec.~\ref{sec:appearance}). 
The method enables animation (Sec.~\ref{sec:animation}), garment transfer and stacking (Sec.~\ref{sec:transfer}), and simulation-ready mesh extraction (Sec.~\ref{sec:meshing}).

\vspace{1mm}
\noindent \textbf{Notation.} 
We index Gaussians by $i$, semantic layers by $l$, and the vertices of a SMPL-X face by $k \in \{1,2,3\}$.

\vspace{-1mm}
\subsection{DAMA Reconstruction}
\label{sec:pipeline}
\vspace{-1mm}
\subsubsection{Coarse Reconstruction from Segmentation}
\label{sec:segmentation}


\noindent \textbf{SMPL-X Gaussians.}
We subdivide and convert the SMPL-X mesh into a Gaussian set $\mathcal{G}^{\text{smplx}}=\{g_i^{\text{smplx}}\}_{i=1}^{N_{\text{faces}}}$ that serves as a body reference during reconstruction. Each Gaussian $g_i^{\text{smplx}}=
(\boldsymbol{\mu}_i^{\text{smplx}},\mathbf{s}_i^{\text{smplx}},\mathbf{q}_i^{\text{smplx}},\alpha_i^{\text{smplx}},\mathbf{c}_i^{\text{smplx}})$ corresponds to one SMPL-X face. The mean $\boldsymbol{\mu}_i$ is placed at the face center, the orientation $\mathbf{q}_i$ is derived from the face plane, and the scale $\mathbf{s}_i$ is set to cover the face area. The color $\mathbf{c}_i$ is initialized as the average skin color estimated from the skin masks, and the opacity $\alpha_i$ is fixed to $1$.

\begin{figure}[t]
\centering
\vspace{-3mm}
\includegraphics[width=.9\columnwidth]{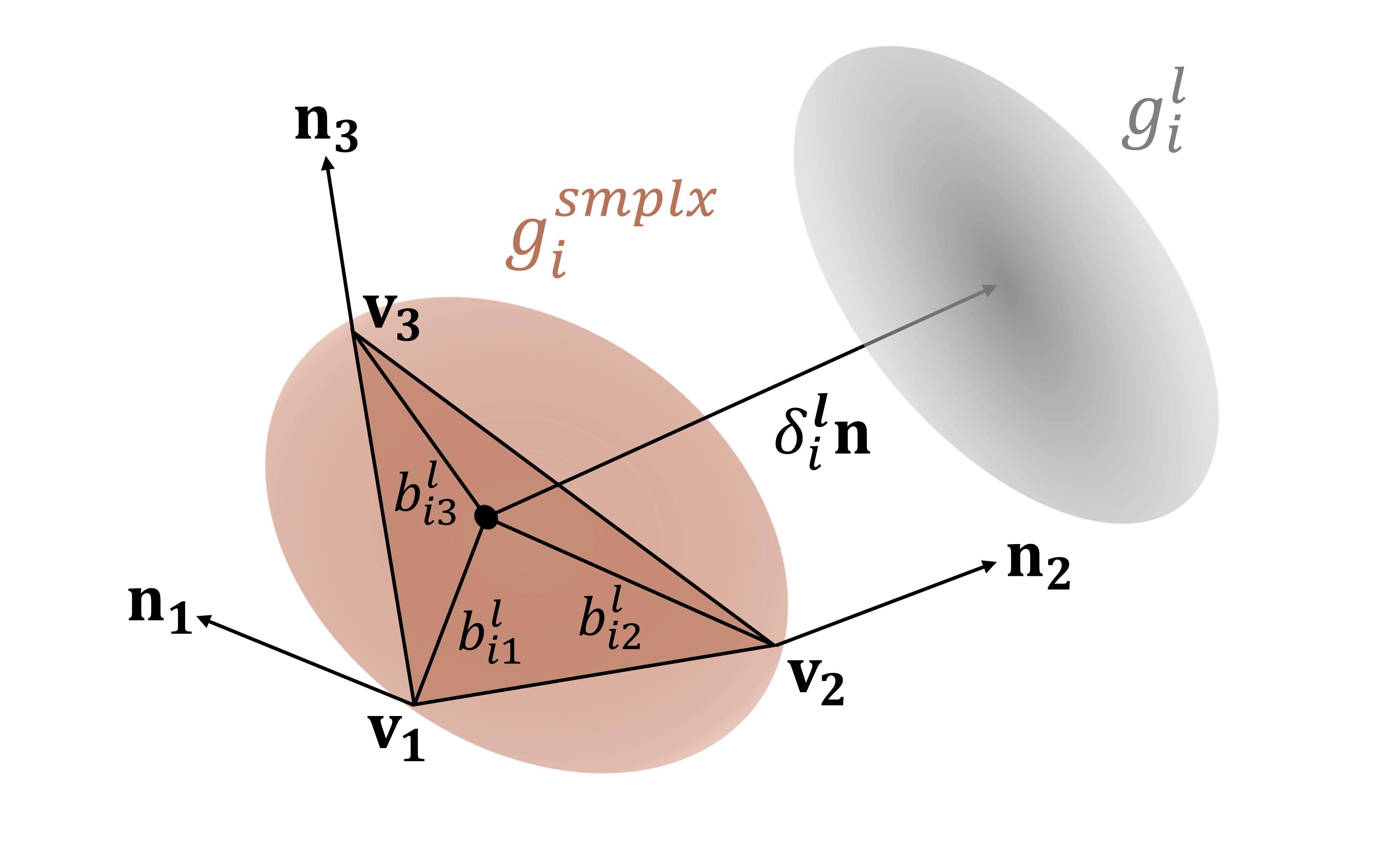}
\vspace{-5mm}
\caption{\textbf{Anchored Gaussian Representation.} The Gaussian mean is expressed using barycentric coordinates on the SMPL-X face and a positive offset along the interpolated normal.}
\vspace{-3mm}
\label{fig:representation}
\end{figure}

\vspace{1mm}
\noindent \textbf{Anchored Gaussian Representation.} 
DAMA represents the avatar layers as sets of Gaussians anchored to the SMPL-X mesh. Our anchoring enables SMPL-X control for reposing, keeps Gaussians close to their corresponding face, prevents lateral drift across the mesh, and ensures layers remain outside the body. We re-parameterize the Gaussian mean into an in-plane position on the SMPL-X face and a positive offset along the normal. Let $\mathbf{v}_k$ and $\mathbf{n}_k$ denote the vertices and vertex normals of the face. For each Gaussian $g_i^l$ in layer $l$, we represent the in-plane position using barycentric coordinates $\mathbf{b}_i^l=(b_{i1}^l,b_{i2}^l,b_{i3}^l)$ with $b_{ik}^l \ge 0$ and $\sum_{k=1}^{3} b_{ik}^l = 1$. The Gaussian mean becomes
\vspace{-3mm}
\begin{equation}
\boldsymbol{\mu}_i^{l} =
\sum_{k=1}^{3} b_{ik}^{l} \mathbf{v}_k +
\delta_i^{l} \sum_{k=1}^{3} b_{ik}^{l} \mathbf{n}_k
\vspace{-3mm}
\end{equation}
where $\delta_i^{l} > 0$ moves the Gaussian along the normal and prevents intersections with the body.  We express the Gaussian orientation $\mathbf{q}^l_i$ relative to the orientation of the corresponding SMPL-X Gaussian $\mathbf{q}^{\text{smplx}}_i$. Let $\mathbf{q}_{r,i}^{l}$ denote the relative rotation; the final orientation becomes $\mathbf{q}_i^{l} = \mathbf{q}^{\text{smplx}}_i \circ \mathbf{q}_{r,i}^{l}$. The optimized variables are the barycentric coordinates $\mathbf{b}_i^{l}$, the offset $\delta_i^{l}$, and the relative rotation $\mathbf{q}_{r,i}^{l}$. Fig.~\ref{fig:representation} illustrates the anchored Gaussian parameterization.

\vspace{1mm}
\noindent \textbf{Segmentation Lifting.}
In this stage, we represent the clothed human as a single segmentation layer
$\mathcal{G}^{\text{seg}}=\{g_i^{\text{seg}}\}_{i=1}^{N_{\text{seg}}}$ anchored to the SMPL-X surface, where
$g_i^{\text{seg}}=(\mathbf{b}_i^{\text{seg}},\delta_i^{\text{seg}},\mathbf{s}_i^{\text{seg}},\mathbf{q}_{r,i}^{\text{seg}},\alpha_i^{\text{seg}},\ell_i)$, where $\mathbf{b}_i$ are barycentric coordinates, $\delta_i$ the normal offset,
$\mathbf{s}_i$ the scale, $\mathbf{q}_{r,i}$ the relative orientation,
$\alpha_i$ the opacity, and $\ell_i$ the semantic label. We initialize $\mathcal{G}^{\text{seg}}$ from $\mathcal{G}^{\text{smplx}}$ by setting $\mathbf{b}_i=(\tfrac{1}{3},\tfrac{1}{3},\tfrac{1}{3})$, $\delta_i$ to a value close to zero, and $\mathbf{q}_{r,i}$ to identity. We copy the scales from $\mathcal{G}^{\text{smplx}}$, fix $\alpha_i=1$, and initialize all labels as skin. We optimize $(\mathbf{b}_i,\delta_i,\mathbf{q}_{r,i},\mathbf{s}_i,\ell_i)$ using a segmentation loss:
\vspace{-1mm}
\begin{equation}
\mathcal{L}_{\text{seg}} =
\lambda_c \mathcal{L}_c +
\lambda_s \mathcal{L}_s +
\lambda_n \mathcal{L}_n +
\lambda_{\ell} \mathcal{L}_{\ell}
\vspace{-2mm}
\end{equation}
$\mathcal{L}_c$ is a photometric loss on rendered segmentation masks following D3GA~\cite{zielonka2025drivable}. Each label is assigned a color and the rendered masks are compared to ground-truth masks. $\mathcal{L}_s$ keeps the scales of $\mathcal{G}^{\text{seg}}$ close to the scales of the corresponding SMPL-X Gaussians to prevent collapse or excessive growth and ensure each Gaussian covers a similar surface area. $\mathcal{L}_n$ aligns Gaussian normals with normals estimated from rendered depth maps following 2DGS~\cite{huang20242dgs}. $\mathcal{L}_{\ell}$ encourages neighboring Gaussians to share similar labels. Disco4D \cite{pang2025disco4d} recomputes nearest neighbors every iteration, while our anchored representation keeps Gaussians near their bound SMPL-X face, allowing neighbors to be precomputed once before optimization, reducing runtime.

To stabilize supervision under alpha compositing, we render $\mathcal{G}^{\text{smplx}}$ together with $\mathcal{G}^{\text{seg}}$ but keep it fixed. 
This forces foreground Gaussians to explain the visible pixels instead of relying on colors from distant Gaussians within the same layer. We randomize the color of $\mathcal{G}^{\text{smplx}}$ each iteration to prevent $\mathcal{G}^{\text{seg}}$ from fitting the body color.

\subsubsection{Topology-Aware Layer Refinement}
\label{sec:refinement}

\vspace{-1mm}
Labels lifted in Stage~1 can be noisy in detailed or self-occluded regions (e.g. hands, face, neck, underarms, inner thighs). We correct these artifacts using SMPL-X mesh topology. Since segmentation Gaussians $\mathcal{G}^{\text{seg}}$ remain aligned with SMPL-X faces (no lateral drift or body intersections), we project their labels to the SMPL-X mesh by assigning each face the label of its bound Gaussian. We then compute connected components of faces sharing the same label, where spurious components below an area threshold are relabeled using the majority label of neighboring faces. This refinement is repeated until no small components remain. The refined face labels are projected back to the bound Gaussians $\mathcal{G}^{\text{seg}}$, producing a cleaned segmentation layer for separation into semantic layers.

\subsubsection{Fine Geometry and Appearance Optimization}
\label{sec:appearance}
\vspace{-1mm}
After refinement, we split the segmentation set $\mathcal{G}^{\text{seg}}$ into semantic layer subsets (e.g., skin, hair, upper, lower, outer). Each layer is represented as
$\mathcal{G}^{l}=\{g_i^{l}\}_{i=1}^{N_l}$,
where
$g_i^{l}=(\mathbf{b}_i^{l},\delta_i^{l},\mathbf{s}_i^{l},\mathbf{q}_{r,i}^{l},\alpha_i^{l},\mathbf{c}_i^{l})$
preserves the anchored parameterization and $\mathbf{c}_i$ denotes the RGB color. In Stage~1, each SMPL-X face had a single Gaussian constrained to roughly cover the face area to obtain coarse geometry. At this stage, we duplicate these Gaussians so that multiple Gaussians attach to each face. We initialize $(\mathbf{b}_i,\delta_i,\mathbf{q}_{r,i})$ from Stage~1, set $\mathbf{c}_i$ to the average color of the masked RGB images for that layer, and fix $\alpha_i=1$. The scales $\mathbf{s}_i$ are initialized small with isotropic scale to capture fine geometry. We then optimize $(\mathbf{b}_i,\delta_i,\mathbf{q}_{r,i},\mathbf{s}_i,\mathbf{c}_i)$ using an appearance loss:
\vspace{-2mm}
\begin{equation}
\mathcal{L}_{\text{app}} =
\lambda_c \mathcal{L}_c +
\lambda_m \mathcal{L}_m +
\lambda_a \mathcal{L}_a +
\lambda_n \mathcal{L}_n +
\lambda_d \mathcal{L}_d +
\lambda_r \mathcal{L}_r
\vspace{-2mm}
\end{equation}
$\mathcal{L}_c$ is a color loss between the rendered RGB image and the masked RGB image of the layer.
$\mathcal{L}_m$ is an $L_1$ loss between the rendered layer mask and the ground-truth mask.
$\mathcal{L}_a$ encourages isotropic scales and prevents collapse or excessive growth.
$\mathcal{L}_n$ is the 2DGS normal loss used in Stage~1.
$\mathcal{L}_d$ and $\mathcal{L}_r$ are canonical distance and rotation losses computed in canonical space that keep Gaussians close to the SMPL-X surface and aligned with the face orientation, stabilizing optimization in occluded or weakly supervised regions.

Prior work \cite{pang2025disco4d,kim2024gala,lin2024layga,chen2026gaussianwardrobe} renders all layers jointly and evaluates losses on the full image, which can cause color leakage between layers. We instead optimize $\mathcal{L}_{\text{app}}$ for each layer independently using masked RGB supervision. Afterward, we render all layers jointly and refine the Gaussian means by optimizing only $(\mathbf{b}_i, \delta_i)$ with $\mathcal{L}_c$ and $\mathcal{L}_m$, while keeping color, scale, and rotation fixed. This step ensures that the layers combine consistently to reconstruct the full avatar. As in Stage~1, we render the fixed body Gaussians $\mathcal{G}^{\text{smplx}}$ with each layer and randomize their color to stabilize alpha compositing and avoid fitting the body color.

We compose the body as $\mathcal{G}^{\text{body}}=\mathcal{G}^{\text{skin}}\cup\mathcal{G}^{\text{hair}}\cup\tilde{\mathcal{G}}^{\text{smplx}}$. Here, $\tilde{\mathcal{G}}^{\text{smplx}}$ contains only SMPL-X Gaussians whose faces are not labeled as garments. These Gaussians complete body regions occluded by clothing. The full avatar $\mathcal{G}^{\text{full}}$ is obtained by taking the union of $\mathcal{G}^{\text{body}}$ and all garment layers.

\subsection{Avatar Animation}
\label{sec:animation}
\vspace{-1mm}
DAMA supports SMPL-X–driven animation through the anchored parameterization. Given a new pose, we deform the SMPL-X mesh using LBS and convert it to Gaussians $\mathcal{G}^{\text{smplx}}$. Each Gaussian $g^{l}_{i}$ in every semantic layer $l$ (skin, hair, or garment) is then updated using its parameters $(\mathbf{b}_i, \delta_i, \mathbf{q}_{r,i})$ with respect to the posed SMPL-X mesh.

\begin{figure*}[t]
    \centering
    \includegraphics[width=\textwidth]{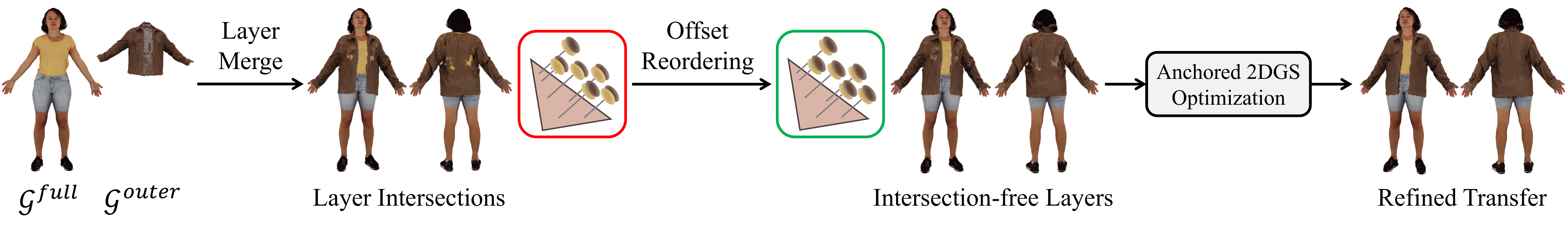}
    \vspace{-5mm}
    \caption{\textbf{Garment Transfer and Stacking.} We transfer a garment layer (outer garment here) to a target avatar by recomputing its Gaussian parameters on the target SMPL-X mesh and merging it with the avatar layers. The naive merge creates intersections. Our representation resolves them by reordering layers and shifting the garment outward using the offsets of lower layers. This offset may distort appearance. We therefore refine the transferred garment alone with anchored 2DGS optimization supervised by its standalone rendering.}
    \label{fig:transfer}
    \vspace{-5mm}
\end{figure*}

\subsection{Garment Transfer and Stacking}
\label{sec:transfer}
\vspace{-1mm}
DAMA enables garment transfer and stacking through the anchored representation. Since Gaussians are bound to SMPL-X faces, we transfer a garment layer $l$ by recomputing its Gaussian means and orientations using its parameters $(\mathbf{b}_i, \delta_i, \mathbf{q}_{r,i})$ and the target SMPL-X mesh. The transferred garment may overlap with existing layers. We resolve these intersections by enforcing a user-defined layer order. We offset layer $l$ by updating its Gaussian means as
\vspace{-2mm}
\begin{equation}
\boldsymbol{\mu}_i^{l} =
\sum_{k=1}^{3} b_{ik}^{l} \mathbf{v}_k +
(\delta_i^{l} + \delta_{\text{prev},i}^{l})
\sum_{k=1}^{3} b_{ik}^{l} \mathbf{n}_k
\vspace{-2mm}
\end{equation}
where $\delta_{\text{prev},i}^{l}$ is the maximum offset of all layers below $l$. The offset may slightly misalign the appearance. We correct it with anchored 2DGS optimization on the transferred garment only. We render all layers and optimize only $(\mathbf{b}_i,\delta_i)$ of the transferred layer with $\mathcal{L}_c$ and $\mathcal{L}_m$. We compute $\mathcal{L}_c$ against the standalone RGB rendering of the transferred garment and $\mathcal{L}_m$ from a rendered mask with the garment in white and other layers in black. Fig.~\ref{fig:transfer} illustrates garment transfer, offset reordering, and the subsequent refinement.

\subsection{Simulation-Ready Mesh Extraction}
\label{sec:meshing}
\vspace{-1mm}
DAMA enables fast conversion of garment Gaussians to a simulation-ready mesh without SDFs or marching cubes. Each Gaussian remains bound to a SMPL-X face, preventing surface drift and body penetration, allowing reuse of SMPL-X connectivity. For each SMPL-X vertex in the garment layer, its position is set to the average of Gaussian means attahced to its incident faces. The garment mesh is formed from the corresponding face subset and Laplacian smoothing is applied. All Gaussians lie outside the body and respect layer ordering, ensuring an intersection-free mesh suitable for cloth simulation.

\label{sec:method}
\section{Experiments and Results}

\vspace{-1mm}
\noindent \textbf{Dataset.}
We evaluate DAMA on all 82 scans of 4D-DRESS \cite{wang20244ddress}. For each sequence, we use the first frame, render 20 circular views (RGB and masks) at $1024\times1024$, and reconstruct the disentangled avatar.

\vspace{1mm}
\noindent \textbf{Baselines.}
We compare with GALA~\cite{kim2024gala} and Disco4D~\cite{pang2025disco4d}, which reconstruct disentangled avatars from single-frame inputs. GALA takes a 3D scan, renders views, segments them with SAM~\cite{kirillov2023segment}, and reconstructs one garment against the body. Disco4D takes a single image, synthesizes multi-view images with diffusion~\cite{blattmann2023stable}, segments them with SegFormer~\cite{xie2021segformer}, and jointly reconstructs the body and garments. For a fair comparison, we adapt both methods to use the same inputs as ours: ground-truth multi-view images, segmentation masks, and SMPL-X fits from 4D-DRESS. GALA additionally requires the 3D scan as input. We also include a standard 2DGS~\cite{huang20242dgs} reconstruction baseline trained on the input images without disentanglement.

\vspace{1mm}
\noindent \textbf{Metrics.}
We evaluate visual quality with PSNR and LPIPS on 12 circular novel views, geometry with two-way Chamfer distance (mm), and physical plausibility with body penetration rate (percentage of intersecting primitives) and average penetration depth (mm). Intersections are computed against the SMPL-X body for DAMA and Disco4D, and against the reconstructed body mesh for GALA.

\subsection{Main Comparisons}
\begin{figure*}[t]
    \centering
    \includegraphics[width=\textwidth]{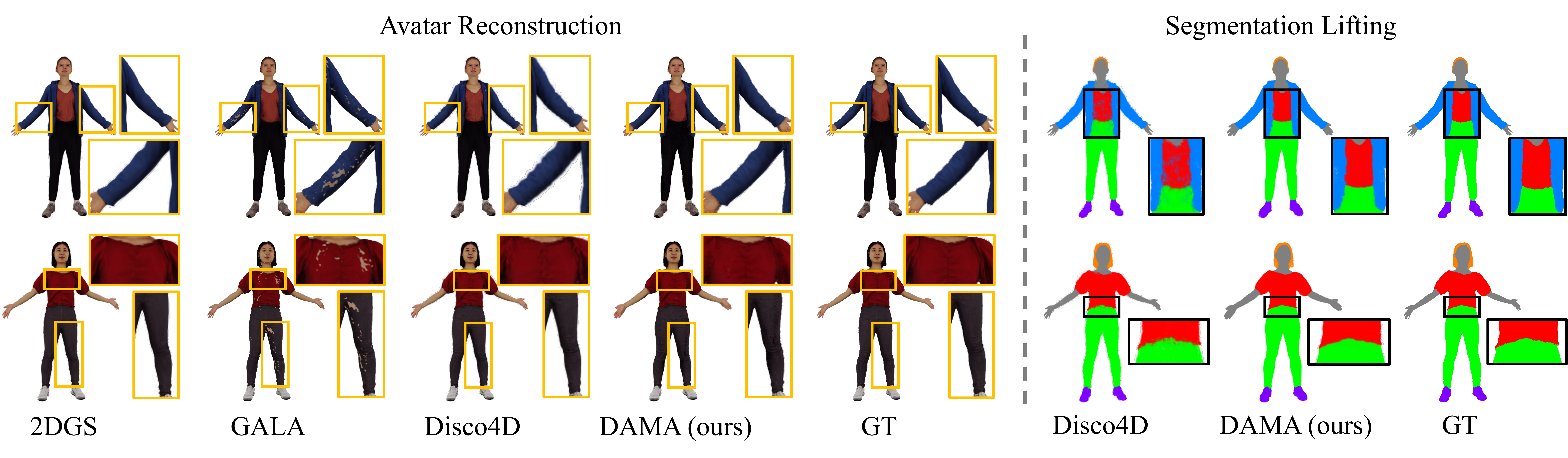}
    \vspace{-7mm}
    \caption{\textbf{Full-Avatar Reconstruction.} GALA shows artifacts from garment–body mesh intersections (left). Disco4D produces noisy boundaries and incorrect lifted regions (right). DAMA reconstructs non-intersecting layered garments with accurate labels.}
    \label{fig:reconstruction}
    \vspace{-0mm}
\end{figure*}

\vspace{-1mm}
\noindent \textbf{Full-Avatar Reconstruction.}
Tab.~\ref{tab:reconstruction} reports full-avatar metrics. DAMA achieves the best geometry (lowest Chamfer distance) and reduces body–garment intersections by over an order of magnitude, supporting physically plausible layering enabled by our representation. We observe slightly lower PSNR, which we attribute to two geometric constraints: Gaussians are restricted to remain outside the body, preventing interior alpha-compositing, and small SMPL-X misalignments (e.g., around fingers) cannot be compensated by placing Gaussians inside the body. Both factors impact pixel-wise metrics such as PSNR. Fig.~\ref{fig:reconstruction} shows qualitative comparisons: GALA exhibits artifacts from garment–body mesh intersections, while Disco4D produces noisy segmentation boundaries and incorrect regions during lifting.

\begin{table}[t]
\vspace{-0mm}
\centering
\small
\setlength{\tabcolsep}{3.5pt}
\begin{tabular*}{\columnwidth}{@{\extracolsep{\fill}}lccccc}
\toprule
& \multicolumn{2}{c}{Appearance} & Geometry & \multicolumn{2}{c}{Penetration} \\
\cmidrule(lr){2-3} \cmidrule(lr){4-4} \cmidrule(lr){5-6}
\multirow{2}{*}{Method} & \multirow{2}{*}{PSNR$\uparrow$} & \multirow{2}{*}{LPIPS$\downarrow$} & CD$\downarrow$ & Rate$\downarrow$ & Depth$\downarrow$ \\
 &  &  & (mm) & (\%) & (mm) \\
\midrule
GALA~\cite{kim2024gala} & 32.43 & \textbf{0.025} & 28.78 & 36.48 & 28.43 \\
Disco4D~\cite{pang2025disco4d} & 32.81 & 0.040 & 28.86 & 49.08 & 18.20 \\
2DGS~\cite{huang20242dgs} & \textbf{35.21} & 0.031 & 21.88 & 51.83 & 8.91 \\
\midrule
DAMA (Ours) & 30.15 & 0.035 & \textbf{19.88} & \textbf{1.46} & \textbf{0.32} \\
\bottomrule
\end{tabular*}
\vspace{-2mm}
\caption{\textbf{Full-Avatar Reconstruction Metrics.} DAMA yields best geometry, minimal intersections, and comparable appearance.}
\vspace{-5mm}
\label{tab:reconstruction}
\end{table}

\begin{figure*}[t]
    \centering
    \vspace{-3mm}
    \includegraphics[width=\textwidth]{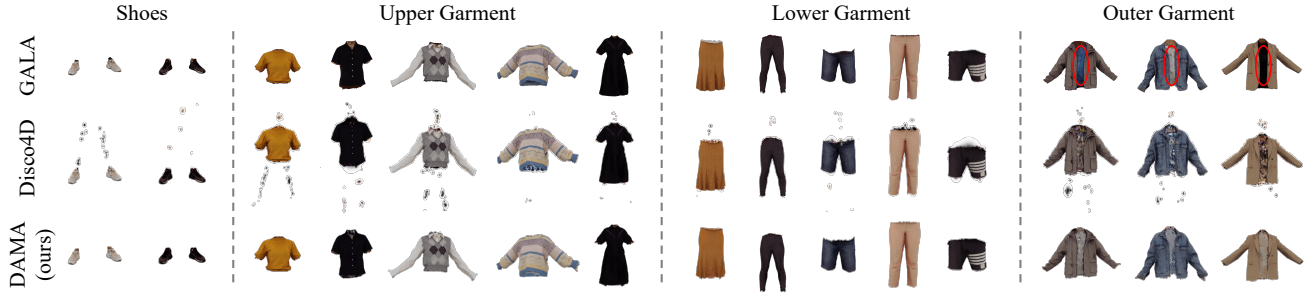}
    \vspace{-7mm}
    \caption{\textbf{Garment Disentanglement.} Disco4D produces noisy labels that appear as floating Gaussians. GALA captures inner garments when extracting the outer layer (red ellipses). DAMA yields cleanly isolated garments through topology-aware refinement.}
    \label{fig:disentanglement}
    \vspace{-5mm}
\end{figure*}

\vspace{1mm}
\noindent \textbf{Garment Disentanglement.}
We evaluate disentanglement using Chamfer distance and penetration metrics for each garment. Tab.~\ref{tab:disentanglement} reports results for upper, lower, and outer garments; Fig.~\ref{fig:disentanglement} shows qualitative comparisons. Both baselines produce ambiguous garment separation. GALA labels mesh faces by mask overlap, ignoring depth ordering, which often selects inner and outer garments together. Disco4D lifts 2D segmentations to 3D using rendered label supervision, which frequently produces small mislabeled regions that appear as floating Gaussians when garments are visualized separately. Our anchored representation and topology-aware refinement (Sec.~\ref{sec:refinement}) avoid these artifacts and yield clean garment separation.

\begin{table}[t]
\centering
\resizebox{\columnwidth}{!}{
\begin{tabular}{llccc}
\toprule
& & \multicolumn{1}{c}{Geometry} & \multicolumn{2}{c}{Penetration} \\
\cmidrule(lr){3-3} \cmidrule(lr){4-5}
Garment & Method & CD (mm)$\downarrow$ & Rate (\%)$\downarrow$ & Depth (mm)$\downarrow$ \\
\midrule
\multirow{3}{*}{Upper}
& GALA~\cite{kim2024gala} & 23.97 & 29.95 & 21.37 \\
& Disco4D~\cite{pang2025disco4d} & 40.31 & 45.20 & 22.88 \\
& DAMA (Ours) & \textbf{23.02} & \textbf{0.56} & \textbf{0.30} \\

\midrule
\multirow{3}{*}{Lower}
& GALA~\cite{kim2024gala} & 23.20 & 35.40 & 20.04 \\
& Disco4D~\cite{pang2025disco4d} & 37.99 & 53.84 & 22.65 \\
& DAMA (Ours) & \textbf{23.09} & \textbf{0.84} & \textbf{0.29} \\

\midrule
\multirow{3}{*}{Outer}
& GALA~\cite{kim2024gala} & 39.42 & 41.01 & 34.00 \\
& Disco4D~\cite{pang2025disco4d} & 31.45 & 29.88 & 19.98 \\
& DAMA (Ours) & \textbf{29.66} & \textbf{0.79} & \textbf{0.35} \\

\bottomrule
\end{tabular}
}
\vspace{-2mm}
\caption{\textbf{Garment Disentanglement Metrics.} DAMA achieves clean garment separation with minimal penetration.}
\vspace{-5mm}
\label{tab:disentanglement}
\end{table}

\vspace{-1mm}
\subsection{Ablations}

\begin{figure}[t]
    \centering
    \vspace{-2mm}
    \includegraphics[width=\columnwidth]{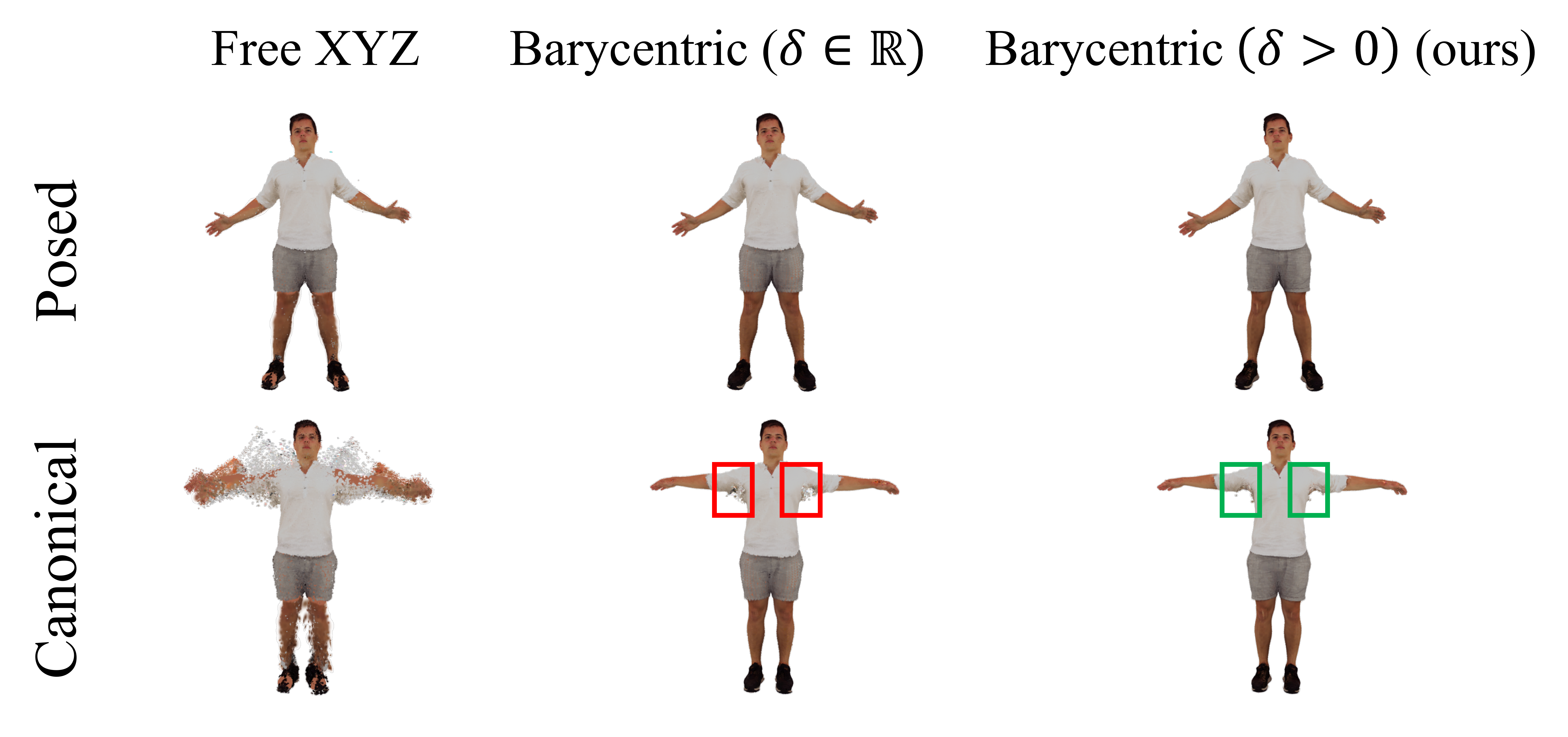}
    \vspace{-6mm}
    \caption{\textbf{Qualitative Ablation of our Gaussian Representation.} Free XYZ causes drifting Gaussians, barycentric with unsigned offset ($\delta \in \mathbb{R}$) produces artifacts, while our positive offset ($\delta > 0$) keeps Gaussians surface-aligned and stable under animation.}
    \label{fig:representation_ablation}
    \vspace{-3mm}
\end{figure}

\vspace{-1mm}
\noindent \textbf{Representation.}
We compare three parameterizations of the Gaussian mean: free XYZ optimization, barycentric coordinates with an unsigned offset ($\delta \in \mathbb{R}$), and barycentric coordinates with a positive offset ($\delta > 0$) (ours). Tab.~\ref{tab:representation_ablation} reports Chamfer distance and penetration metrics. Free XYZ and $\delta \in \mathbb{R}$ produce intersections, while our $\delta > 0$ achieves comparable Chamfer distance with significantly lower  penetration. Fig.~\ref{fig:representation_ablation} shows results in canonical and posed space. Free XYZ causes drifting and floating Gaussians in animation, while $\delta \in \mathbb{R}$ produces artifacts in weakly supervised regions. Our $\delta > 0$ parameterization keeps Gaussians near the surface and stable under animation.

\begin{table}[t]
\centering
\small
\setlength{\tabcolsep}{3.5pt}
\begin{tabular*}{\columnwidth}{@{\extracolsep{\fill}}lccc}
\toprule
& \multicolumn{1}{c}{Geometry} & \multicolumn{2}{c}{Penetration} \\
\cmidrule(lr){2-2} \cmidrule(lr){3-4}
Representation & CD (mm)$\downarrow$ & Rate (\%)$\downarrow$ & Depth (mm)$\downarrow$ \\
\midrule
Free XYZ & 18.05 & 29.38 & 8.45 \\
Bary ($\delta \in \mathbb{R}$) & \textbf{17.91} & 19.68 & 7.63 \\
Bary ($\delta > 0$) (Ours) & 19.88 & \textbf{1.46} & \textbf{0.32} \\
\bottomrule
\end{tabular*}
\vspace{-1mm}
\caption{\textbf{Quantitative Ablation of our Gaussian Representation.} Enforcing a positive offset drastically reduces body penetration with only a small Chamfer distance increase.}
\label{tab:representation_ablation}
\vspace{-3mm}
\end{table}

\begin{table}[t]
\centering
\small
\setlength{\tabcolsep}{3.5pt}
\begin{tabular*}{\columnwidth}{@{\extracolsep{\fill}}lccc}
\toprule
Method & mAcc$\uparrow$ & mIoU$\uparrow$ & mF1$\uparrow$ \\
\midrule
DAMA w/o $\mathcal{L}_\ell$ & 0.9700 & 0.9350 & 0.9648 \\
DAMA w/o refinement & 0.9718 & 0.9377 & 0.9665 \\
DAMA (Ours) & \textbf{0.9719} & \textbf{0.9383} & \textbf{0.9669} \\
\bottomrule
\end{tabular*}
\vspace{-2mm}
\caption{\textbf{Quantitative Segmentation Lifting Ablation.} Our full pipeline achieves the best segmentation metrics.}
\label{tab:segmentation_ablation}
\vspace{-4mm}
\end{table}

\begin{figure*}[t]
    \centering
    \includegraphics[width=\textwidth]{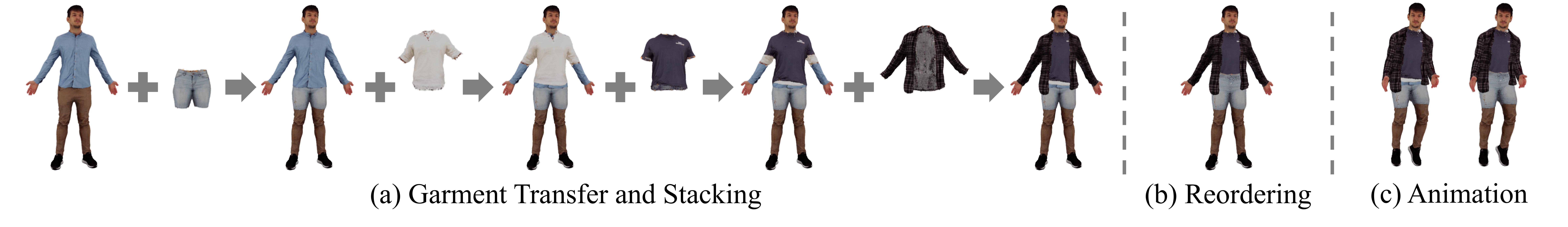}
    \vspace{-8mm}
    \caption{\textbf{Garment Stacking and Reordering.} DAMA enables garment transfer between avatars, garment stacking with collision resolution, reordering of semantic layers, and SMPL-X-driven animation.}
    \label{fig:stacking_and_reordering}
    \vspace{-3mm}
\end{figure*}

\begin{figure*}[t]
    \centering
    \vspace{-2mm}
    \includegraphics[width=\textwidth]{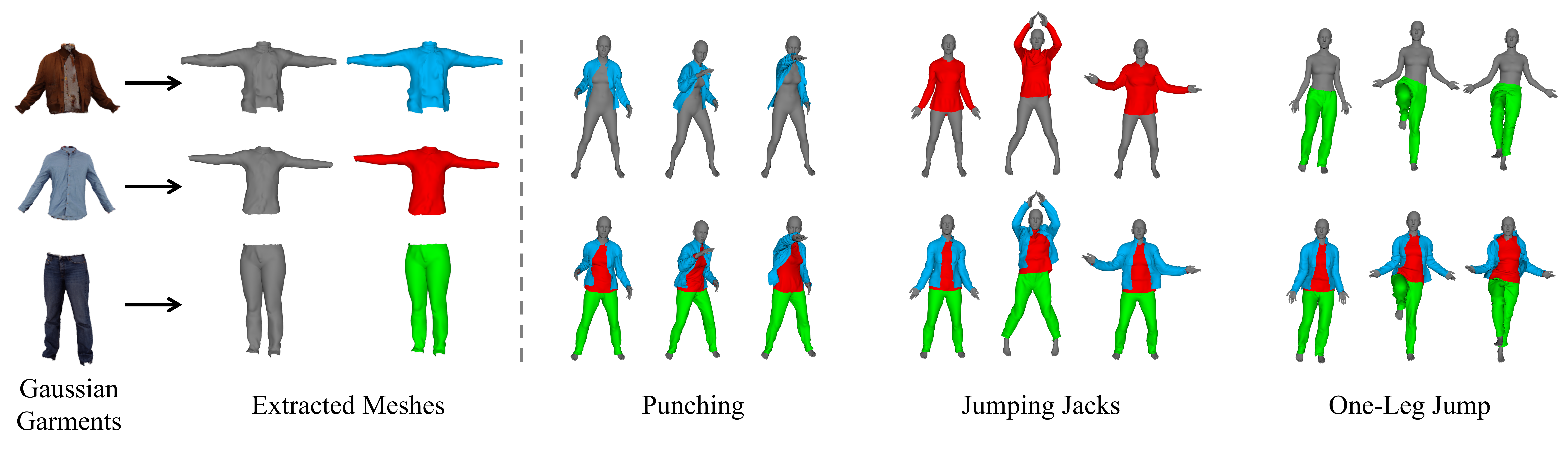}
    \vspace{-8mm}
    \caption{\textbf{Clothing Simulation.} DAMA converts garment geometry to meshes that can be simulated in CLO3D \cite{clo3d}. We show simulation of individual garments (top) and  stacked garments (bottom) driven by SMPL-X animation from AMASS \cite{mahmood2019amass}.}
    \label{fig:simulation}
    \vspace{-4mm}
\end{figure*}

\begin{figure}[h]
    \centering
    \vspace{-2mm}
    \includegraphics[width=\columnwidth]{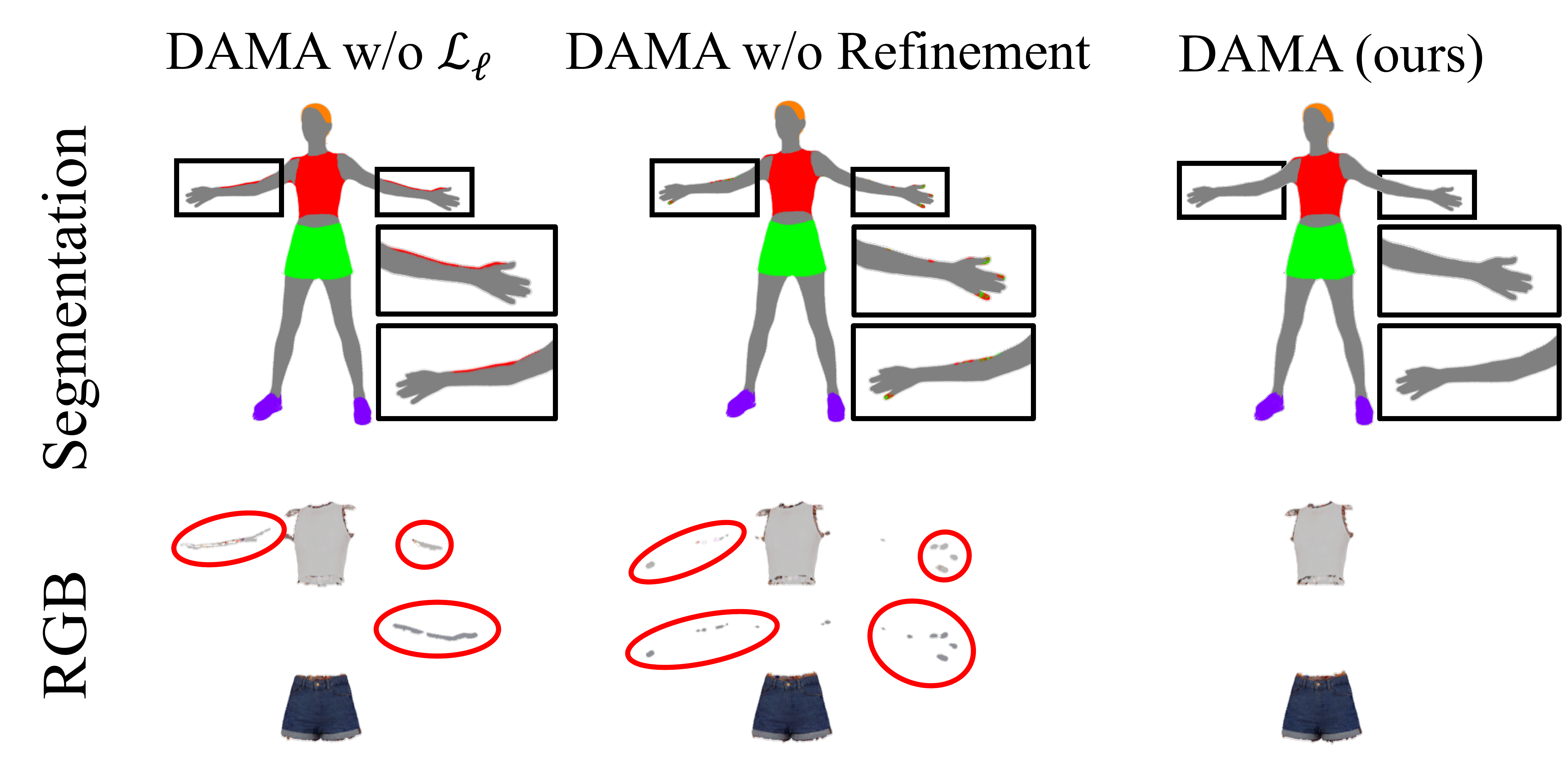}
    \vspace{-6mm}
    \caption{\textbf{Qualitative Segmentation Lifting Ablation.} Removing the smoothness loss produces large incorrect regions, removing topology-based refinement leaves small noisy patches, while our full pipeline yields clean garment separation.}
    \label{fig:segmentation_ablation}
    \vspace{-5mm}
\end{figure}

\vspace{2mm}
\noindent \textbf{Segmentation Lifting Pipeline.}
We ablate the label smoothness loss $\mathcal{L}_\ell$ and the topology-based refinement. Tab.~\ref{tab:segmentation_ablation} reports mAcc, mIoU, and mF1 on rendered masks. Numerical differences are small, however qualitative effects are clear (Fig.~\ref{fig:segmentation_ablation}). Removing smoothness creates large mislabeled regions, while removing refinement leaves small noisy areas from label lifting, which appear as floating Gaussians when garments are visualized separately. Our full pipeline produces accurate labels and clean separation.

\vspace{-0mm}
\subsection{Applications}

\vspace{-1mm}
\noindent \textbf{Garment Stacking and Reordering.}
Our representation enables garment transfer and stacking on existing layers, with collisions resolved by offset ordering (Sec.~\ref{sec:transfer}). Semantic layers can be reordered or reposed with SMPL-X. Fig.~\ref{fig:stacking_and_reordering} shows stacking, reordering, and animation.

\vspace{1mm}
\noindent \textbf{Simulation-Ready Mesh Conversion.}
Body-conforming garments can be quickly converted to meshes (Sec.~\ref{sec:meshing}). The extracted meshes preserve intersection-free layering and can be directly simulated. Fig.~\ref{fig:simulation} shows CLO3D \cite{clo3d} simulations of individual (top) and stacked garments (bottom) using AMASS motion sequences \cite{mahmood2019amass}.

\label{sec:experiments}
\section{Conclusion}

\vspace{-1mm}
We introduced DAMA, a method for reconstructing clothed avatars with physically plausible layering. DAMA anchors Gaussian splats to SMPL-X faces using barycentric coordinates and a strictly positive normal offset. This representation keeps Gaussians tied to the surface, enforces outward layering, and prevents body intersections. The pipeline lifts 2D segmentations, refines labels using SMPL-X topology, and optimizes geometry and appearance for each layer. The anchored formulation enables the refinement step: lifted labels can be projected to the mesh and corrected using mesh connectivity, removing noise from segmentation lifting and producing stable garment boundaries. Evaluated on the full 4D-DRESS dataset, DAMA shows accurate geometry and significantly reduced interpenetration while maintaining photorealistic quality. The representation further enables garment stacking, layer reordering, SMPL-X animation, and fast conversion of garments to simulation-ready meshes. Future work could extend the representation to learn garment deformation from video or support loose clothing animation while preserving explicit layering.

\paragraph{Acknowledgments.} This work is made possible by funding from the Carl Zeiss
Foundation. This work is also funded by the Deutsche Forschungsgemeinschaft (DFG, German Research Foundation) - 409792180 (Emmy Noether Programme, project: Real Virtual Humans) and the German Federal Ministry of Education and Research (BMBF): Tübingen
AI Center, FKZ: 01IS18039A. Daniel Eskandar is supported by the Konrad Zuse School of
Excellence in Learning and Intelligent Systems (ELIZA) through the
DAAD programme Konrad Zuse School of Excellence in Artificial
Intelligence, sponsored by the Federal Ministry of Education and
Research. Berna Kabadayi is supported
by the International Max Planck Research School for Intelligent Systems (IMPRS-IS). Gerard Pons-Moll is a member of the Machine Learning Cluster of
Excellence, EXC number 2064/1 – Project number 390727645.

\label{sec:conclusion}
{
    \small
    \bibliographystyle{ieeenat_fullname}
    \bibliography{main}
}

\clearpage
\maketitlesupplementary

\setcounter{section}{0}
\renewcommand{\thesection}{\Alph{section}}

\section{Implementation Details}

\subsection{Losses}

\noindent \textbf{Color Loss.}
We use an $L_1$ loss between the rendered image and the ground-truth image:
\vspace{-2mm}
\begin{equation}
\mathcal{L}_c = \| I_{\text{rend}} - I_{\text{gt}} \|_1
\vspace{-2mm}
\end{equation}
For segmentation lifting, $I_{\text{rend}}$ contains the rendered label colors assigned to semantic classes.  
For appearance optimization, it contains the rendered RGB colors.

\vspace{2mm}
\noindent \textbf{Scale Loss.}
We keep the scales of segmentation Gaussians close to the scales of the corresponding SMPL-X Gaussians to preserve similar surface coverage.  
Let $\mathbf{s}_i^{\text{seg}}$ denote the scale of Gaussian $g_i^{\text{seg}}$ and $\mathbf{s}_i^{\text{smplx}}$ the scale of its corresponding SMPL-X Gaussian.  
We use an $L_1$ loss:
\vspace{-2mm}
\begin{equation}
\mathcal{L}_s =
\frac{1}{N_{\text{seg}}}
\sum_{i=1}^{N_{\text{seg}}}
\left\| \mathbf{s}_i^{\text{seg}} - \mathbf{s}_i^{\text{smplx}} \right\|_1
\vspace{-2mm}
\end{equation}

\vspace{2mm}
\noindent \textbf{Normal Loss.}
$\mathcal{L}_n$ aligns Gaussian normals with normals estimated from rendered depth maps.  
We use the same formulation and implementation as the normal regularization introduced in 2DGS~\cite{huang20242dgs}.

\vspace{2mm}
\noindent \textbf{Label Smoothness Loss.}
Let $\mathbf{p}_i$ denote the label probability vector of Gaussian $g_i^{\text{seg}}$, with label $\ell^{seg}_i=\arg\max(\mathbf{p}_i)$. 
We encourage neighboring Gaussians to share similar label distributions.  
For precomputed neighbors $\mathcal{N}(i)$ we compute the KL divergence and average over all $N_{\text{seg}}$ Gaussians:
\vspace{-2mm}
\begin{equation}
\mathcal{L}_{\ell} =
\frac{1}{N_{\text{seg}}}
\sum_{i=1}^{N_{\text{seg}}}
\frac{1}{|\mathcal{N}(i)|}
\sum_{j \in \mathcal{N}(i)}
D_{\text{KL}}(\mathbf{p}_i \,\|\, \mathbf{p}_j)
\vspace{-2mm}
\end{equation}

\vspace{2mm}
\noindent \textbf{Mask Loss.}
We use an $L_1$ loss between the rendered layer mask and the ground-truth layer mask:
\vspace{-2mm}
\begin{equation}
\mathcal{L}_m = \| M_{\text{rend}} - M_{\text{gt}} \|_1
\vspace{-2mm}
\end{equation}

\vspace{2mm}
\noindent \textbf{Anisotropic Loss.}
We use the anisotropic regularizer $\mathcal{L}_a$ introduced in PhysGaussian~\cite{xie2024physgaussian}.

\vspace{2mm}
\noindent \textbf{Canonical Distance Loss.}
We use an $L_2$ loss to keep Gaussians close to the SMPL-X surface in canonical space.  
Let $\boldsymbol{\mu}^l_i$ be the Gaussian mean belonging to layer $l$ and $\boldsymbol{\mu}_i^{\text{smplx}}$ the center of its bound SMPL-X face:
\vspace{-2mm}
\begin{equation}
\mathcal{L}_d =
\frac{1}{N_l}
\sum_{i=1}^{N_l}
\|\boldsymbol{\mu}^l_i - \boldsymbol{\mu}_i^{\text{smplx}}\|_2
\vspace{-2mm}
\end{equation}

\vspace{2mm}
\noindent \textbf{Canonical Rotation Loss.}
We align Gaussian orientations with the orientation of their bound SMPL-X face in canonical space.  
Let $\mathbf{q}^l_i$ denote the Gaussian rotation belonging to layer $l$ and $\mathbf{q}_i^{\text{smplx}}$ the SMPL-X face rotation, both in canonical space:
\vspace{-2mm}
\begin{equation}
\mathcal{L}_r =
\frac{1}{N_l}
\sum_{i=1}^{N_l}
\left(1 - \langle \mathbf{q}^l_i , \mathbf{q}_i^{\text{smplx}} \rangle \right)
\vspace{-2mm}
\end{equation}

\subsection{Topology-Aware Label Refinement Algorithm}

We provide the exact algorithm for the topology-aware refinement stage.

After Stage~1, each Gaussian is assigned a label \(\ell_i\). These labels can be noisy. We project the labels onto the SMPL-X mesh by associating each Gaussian with its corresponding face \(f_i\), and refine them on the mesh topology to obtain \(\ell_i^{\mathrm{ref}}\).

Let \(A_i\) denote the area of face \(f_i\). We define a face adjacency graph \(\mathcal{G} = (\mathcal{F}, \mathcal{E})\), where \((f_i, f_j) \in \mathcal{E}\) if the two faces share an edge. The neighbors of a face \(f_i\) are \(\mathcal{N}(i) = \{j \mid (f_i, f_j) \in \mathcal{E}\}\).

We extract connected components \(C \subset \mathcal{F}\) of faces sharing the same label. Let \(A(C) = \sum_{i \in C} A_i\). We introduce an area threshold \(\tau\) and treat components with \(A(C) < \tau\) as spurious, reassigning them to the dominant label of their neighboring faces. This enforces spatial consistency while preserving large regions, yielding refined labels \(\{\ell_i^{\mathrm{ref}}\}\). The full procedure is summarized in Alg.~\ref{alg:refinement}.

\begin{algorithm}[H]
\caption{Topology-Aware Label Refinement}
\label{alg:refinement}
\begin{algorithmic}[1]
\STATE \textbf{Input:} faces \(\{f_i\}\), labels \(\{\ell_i\}\), areas \(\{A_i\}\), threshold \(\tau\)
\STATE Build adjacency graph \(\mathcal{G}\)

\STATE Initialize \(\ell_i^{\mathrm{ref}} \leftarrow \ell_i\)

\REPEAT
    \STATE \algcomment{same-label regions}
    \STATE Extract connected components \(C \subset \mathcal{F}\) such that \(\ell_i^{\mathrm{ref}} = \ell_j^{\mathrm{ref}}\ \forall\, i,j \in C\)

    \FOR{each component \(C\)}
        \STATE \algcomment{compute area}
        \STATE \(A(C) \leftarrow \sum_{i \in C} A_i\)

        \IF{\(A(C) < \tau\)}
            \STATE \algcomment{find neighbors}
            \STATE \(\mathcal{N}(C) \leftarrow \{ j \notin C \mid \exists\, i \in C,\ j \in \mathcal{N}(i) \}\)

            \STATE \algcomment{majority vote}
            \STATE \(\ell^\star \leftarrow \mathrm{mode}\big(\{\ell_j^{\mathrm{ref}} \mid j \in \mathcal{N}(C)\}\big)\)

            \STATE \algcomment{reassign labels}
            \STATE \(\ell_i^{\mathrm{ref}} \leftarrow \ell^\star \quad \forall i \in C\)
        \ENDIF
    \ENDFOR
\UNTIL{no change in \(\ell^{\mathrm{ref}}\)}

\STATE \textbf{Output:} refined labels \(\{\ell_i^{\mathrm{ref}}\}\)
\end{algorithmic}
\end{algorithm}

\subsection{Optimization and Runtime}

We set the loss weights as follows: $\lambda_c{=}1$, $\lambda_s{=}10$, $\lambda_n{=}0.1$, $\lambda_{\ell}{=}0.1$, $\lambda_a{=}100$, $\lambda_d{=}1$, and $\lambda_r{=}100$. All experiments run on a single NVIDIA A100 GPU. In Stage~1, we optimize $\mathcal{G}^{\text{seg}}$ for 10k iterations ($\sim$3 min) and enable the label smoothness loss $\mathcal{L}_{\ell}$ after 5k iterations. In Stage~3, we optimize each semantic layer independently for 2k iterations ($\sim$1.5 min per layer), followed by a final joint optimization of all layers for 2k iterations. The full method takes about 10--15 minutes depending on the number of layers.

\section{Evaluation Metrics}

Let $\mathcal{V}^{\mathrm{gt}} = \{\mathbf{v}_i^{\mathrm{gt}}\}_{i=1}^{N}$ denote the set of ground-truth scan vertices and $\mathcal{V}^{\mathrm{rec}} = \{\mathbf{v}_j^{\mathrm{rec}}\}_{j=1}^{M}$ the set of reconstructed 3D points, represented as Gaussian means for Gaussian-based methods or mesh vertices for mesh-based methods. The body surface is represented by $\mathcal{V}^{\mathrm{body}} = \{\mathbf{v}_k^{\mathrm{body}}\}_{k=1}^{K}$ with corresponding outward normals $\{\mathbf{n}_k\}_{k=1}^{K}$. All quantities are evaluated in the posed space.

\noindent \textbf{Geometric Accuracy.}
Geometric accuracy is quantified using the two-way Chamfer distance:
\begin{equation}
\begin{aligned}
\mathrm{CD} =\;&
\frac{1}{N} \sum_{i=1}^{N} \min_{j} \|\mathbf{v}_i^{\mathrm{gt}} - \mathbf{v}_j^{\mathrm{rec}}\|_2 \\
&+
\frac{1}{M} \sum_{j=1}^{M} \min_{i} \|\mathbf{v}_j^{\mathrm{rec}} - \mathbf{v}_i^{\mathrm{gt}}\|_2 .
\end{aligned}
\end{equation}
capturing bidirectional proximity and ensuring that reconstructed points cover the ground truth points while remaining close to them.

\noindent \textbf{Physical Plausibility.}
Physical plausibility is evaluated via the signed distance $d_j$ of each reconstructed point $\mathbf{v}_j^{\mathrm{rec}} \in \mathcal{V}^{\mathrm{rec}}$ to the body surface:
\begin{equation}
d_j = \min_{k} \; (\mathbf{v}_j^{\mathrm{rec}} - \mathbf{v}_k^{\mathrm{body}}) \cdot \mathbf{n}_k,
\end{equation}
which indicates whether a point lies outside the body or penetrates it along the local surface normal. Penetration depth is defined as:
\begin{equation}
\mathrm{PD} =
\frac{1}{|\{j \mid d_j < 0\}|} \sum_{j:\, d_j < 0} (-d_j),
\end{equation}
capturing the average extent of interpenetration. The penetration rate is given by:
\begin{equation}
\mathrm{PR} = \frac{|\{j \mid d_j < 0\}|}{M},
\end{equation}
reflecting the proportion of reconstructed points that lie inside the body.

\section{Additional Loss Ablations}

\begin{figure}[h]
    \centering
    \includegraphics[width=\columnwidth]{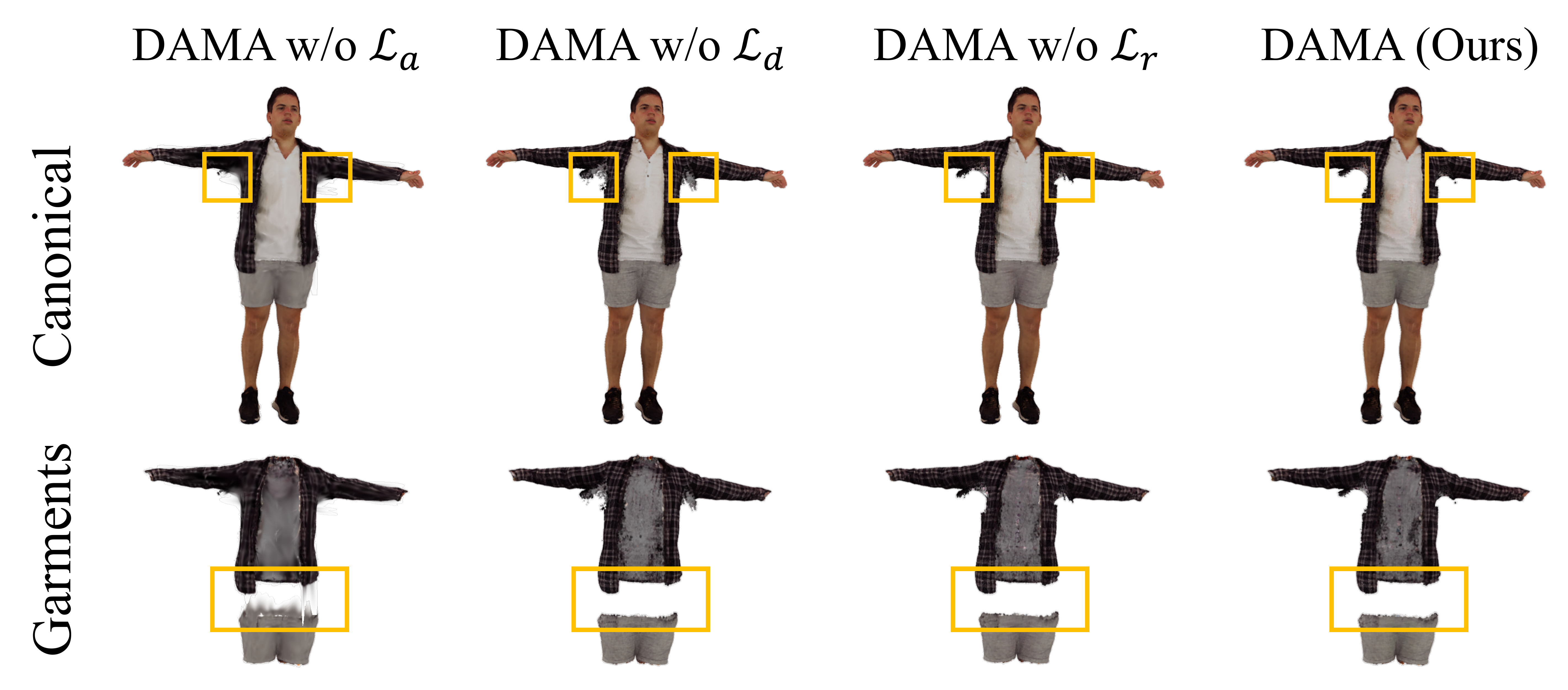}
    \caption{\textbf{Additional Loss Ablations.} Effect of removing $\mathcal{L}_a$, $\mathcal{L}_d$, and $\mathcal{L}_r$.}
    \label{fig:losses_ablation}
\end{figure}

We ablate $\mathcal{L}_a$, $\mathcal{L}_d$, and $\mathcal{L}_r$ to study their individual effects. The loss $\mathcal{L}_a$ prevents Gaussian shrinkage or explosion, while $\mathcal{L}_d$ and $\mathcal{L}_r$ stabilize weakly supervised regions (e.g., underarms), reducing noisy geometry during animation (Fig.~\ref{fig:losses_ablation}).

\section{Additional Applications and Results}

\begin{figure}[h]
    \centering
    \includegraphics[width=\columnwidth]{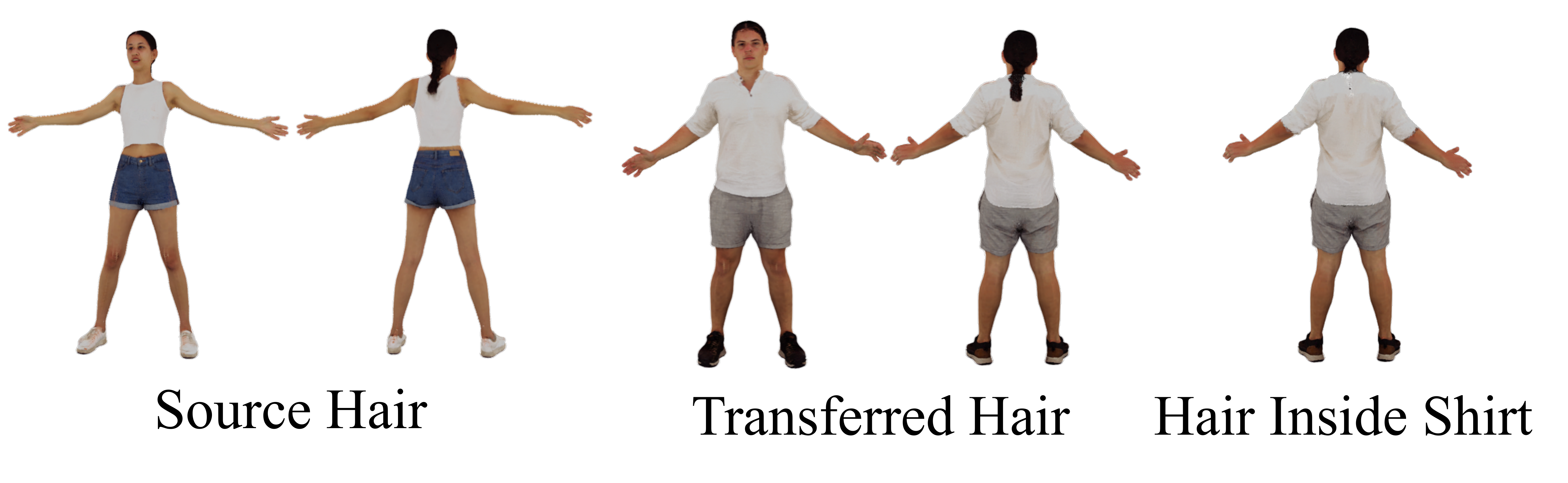}
    \caption{\textbf{Hair Transfer.} Hair transferred from a source subject and reordered.}
    \label{fig:hair}
\end{figure}

\noindent \textbf{Hair Transfer.}
Our representation naturally extends to hair. Fig.~\ref{fig:hair} illustrates transferring hair from a source subject to a target, along with reordering its layer.

\begin{figure*}[t]
    \centering
    \includegraphics[width=\textwidth]{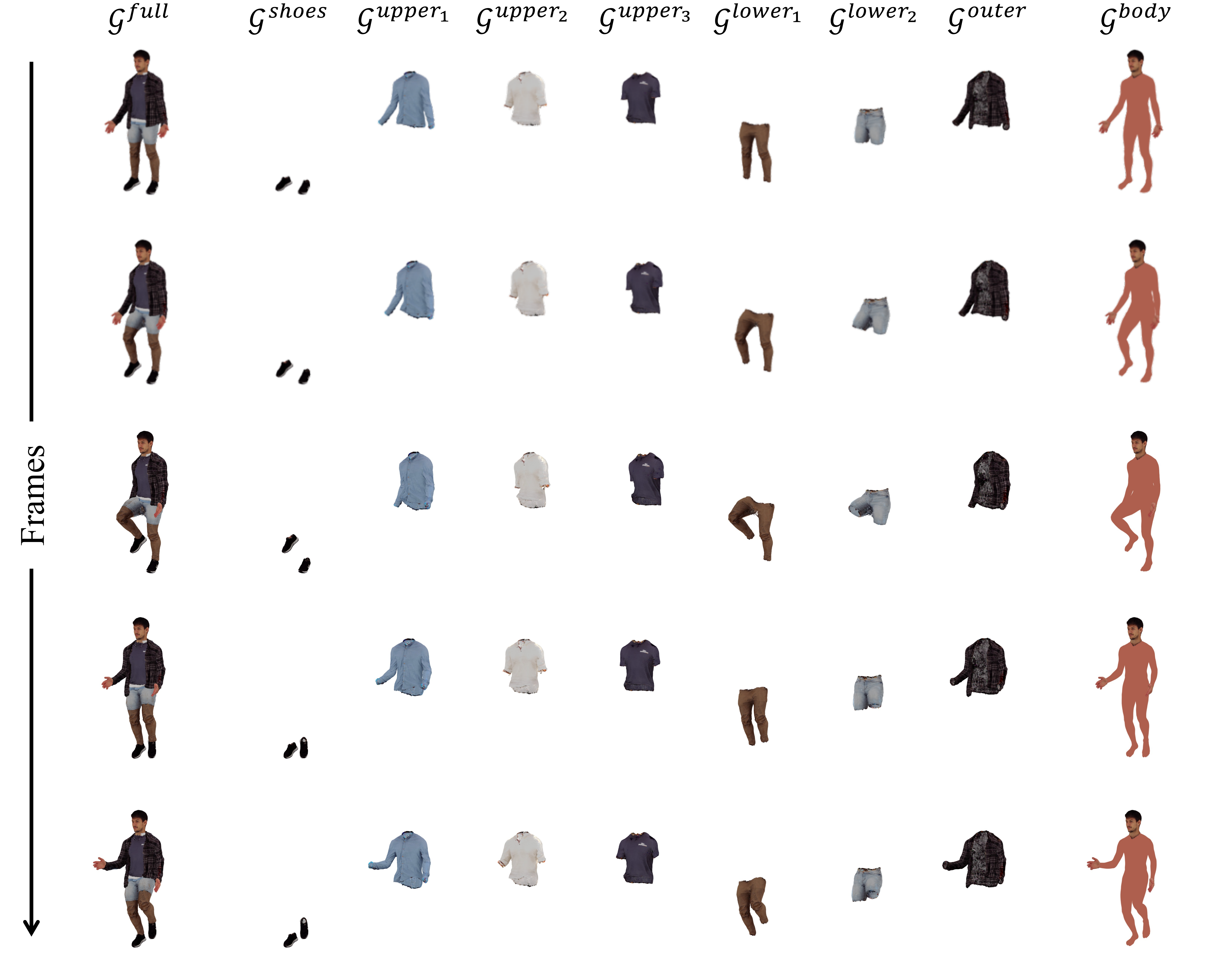}
    \vspace{-8mm}
    \caption{\textbf{SMPL-X–Driven Avatar Animation.}
    We animate the reconstructed avatar with transferred and stacked garments using SMPL-X motion sequences from AMASS \cite{mahmood2019amass}. The sequence shows that the layered garments deform consistently with the body while preserving their ordering and separation throughout the motion.}
    \label{fig:animation2}
    \vspace{-0mm}
\end{figure*}

\begin{figure*}[t]
    \centering
    \vspace{-2mm}
    \includegraphics[width=\textwidth]{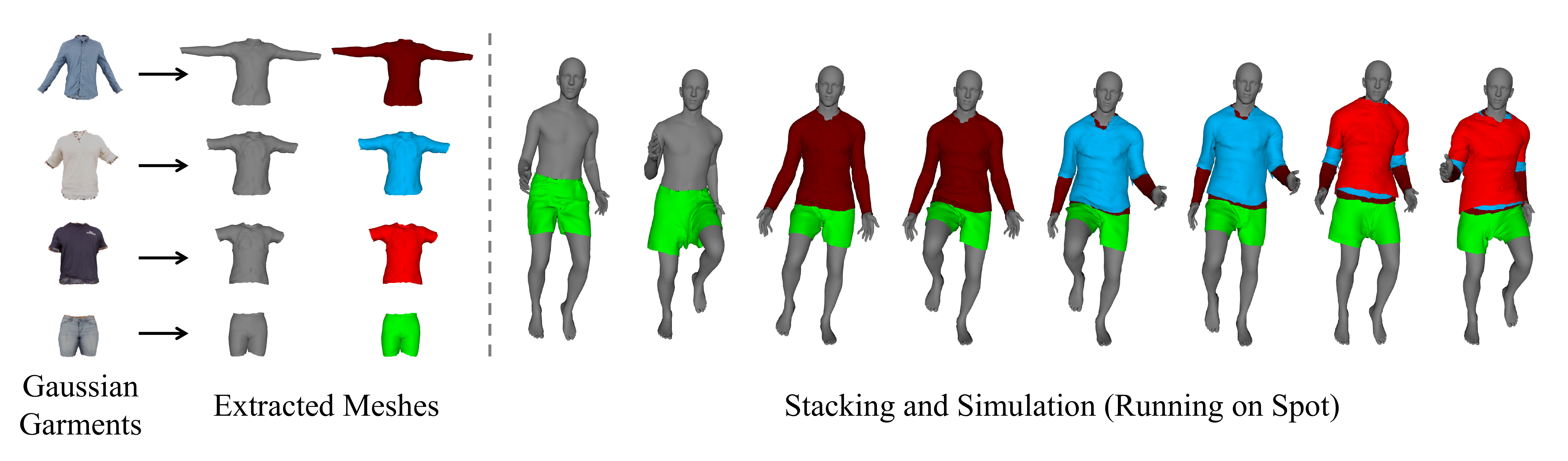}
    \vspace{-7mm}
    \caption{\textbf{Additional Clothing Simulation Example.}
    We show an additional example with one lower garment and three upper garments. 
    (Left) Simulation-ready meshes extracted from the Gaussian layers. 
    (Right) CLO3D~\cite{clo3d} simulation driven by a running-on-spot motion sequence from AMASS \cite{mahmood2019amass}. 
    The garments are progressively stacked, showing that the extracted meshes preserve layer ordering and remain stable during simulation.}
    \vspace{-0mm}
\label{fig:simulation2}
\end{figure*}

\noindent \textbf{Additional Results.}
We further present SMPL-X–driven animation of stacked Gaussian garments with preserved layer ordering (Fig.~\ref{fig:animation2}). We also include additional simulation results of stacked garment meshes extracted from the Gaussians (Fig.~\ref{fig:simulation2}).

\newpage

\end{document}